\documentclass[10pt,twocolumn,letterpaper]{article}

\usepackage[accsupp]{axessibility} 
\usepackage[pagenumbers]{cvpr} 

\usepackage{graphicx}
\usepackage{amsmath}
\usepackage{amssymb}
\usepackage{booktabs}
\usepackage{bm}
\usepackage{multirow}

\usepackage{pifont}
\usepackage{xcolor}         
\definecolor{demphcolor}{RGB}{144,144,144}

\usepackage{colortbl}
\definecolor{mygray}{gray}{0.95}

\usepackage[pagebackref,breaklinks,colorlinks]{hyperref}

\usepackage[capitalize]{cleveref}
\crefname{section}{Sec.}{Secs.}
\Crefname{section}{Section}{Sections}
\Crefname{table}{Table}{Tables}
\crefname{table}{Tab.}{Tabs.}

\def\cvprPaperID{4167} 
\def\confName{CVPR}
\def\confYear{2023}

\begin{document}

\title{Cap4Video: What Can Auxiliary Captions Do for Text-Video Retrieval?}

\author{%
Wenhao Wu$^{1,2}$\thanks{Equal contribution.}\qquad
Haipeng Luo$^{3*}$\qquad
Bo Fang$^{3}$\qquad
Jingdong Wang$^{2}$\qquad
Wanli Ouyang$^{4,1}$\\
$^1$The University of Sydney \qquad $^2$Baidu Inc. \\ 
$^3$University of Chinese Academy of Sciences  \qquad $^4$Shanghai AI Laboratory\\
{\tt\small whwu.ucas@gmail.com}
}

\maketitle

\begin{abstract}

Most existing text-video retrieval methods focus on cross-modal matching between the visual content of videos and textual query sentences. However, in real-world scenarios, online videos are often accompanied by relevant text information such as titles, tags, and even subtitles, which can be utilized to match textual queries. 
This insight has motivated us to propose a novel approach to text-video retrieval, where we directly generate associated captions from videos using zero-shot video captioning with knowledge from web-scale pre-trained models (e.g., CLIP and GPT-2).
Given the generated captions, a natural question arises: what benefits do they bring to text-video retrieval? To answer this, we introduce Cap4Video, a new framework that leverages captions in three ways: i) Input data: video-caption pairs can augment the training data. ii) Intermediate feature interaction: we perform cross-modal feature interaction between the video and caption to produce enhanced video representations. iii) Output score: the Query-Caption matching branch can complement the original Query-Video matching branch for text-video retrieval.
We conduct comprehensive ablation studies to demonstrate the effectiveness of our approach. Without any post-processing, Cap4Video achieves state-of-the-art performance on four standard text-video retrieval benchmarks: MSR-VTT (51.4\%), VATEX (66.6\%), MSVD (51.8\%), and DiDeMo (52.0\%). 
The code is available at \url{https://github.com/whwu95/Cap4Video}.

\end{abstract}

\section{Introduction}
\label{sec:intro}

\begin{figure}[t]
\begin{center}
\includegraphics[width=0.98\columnwidth]{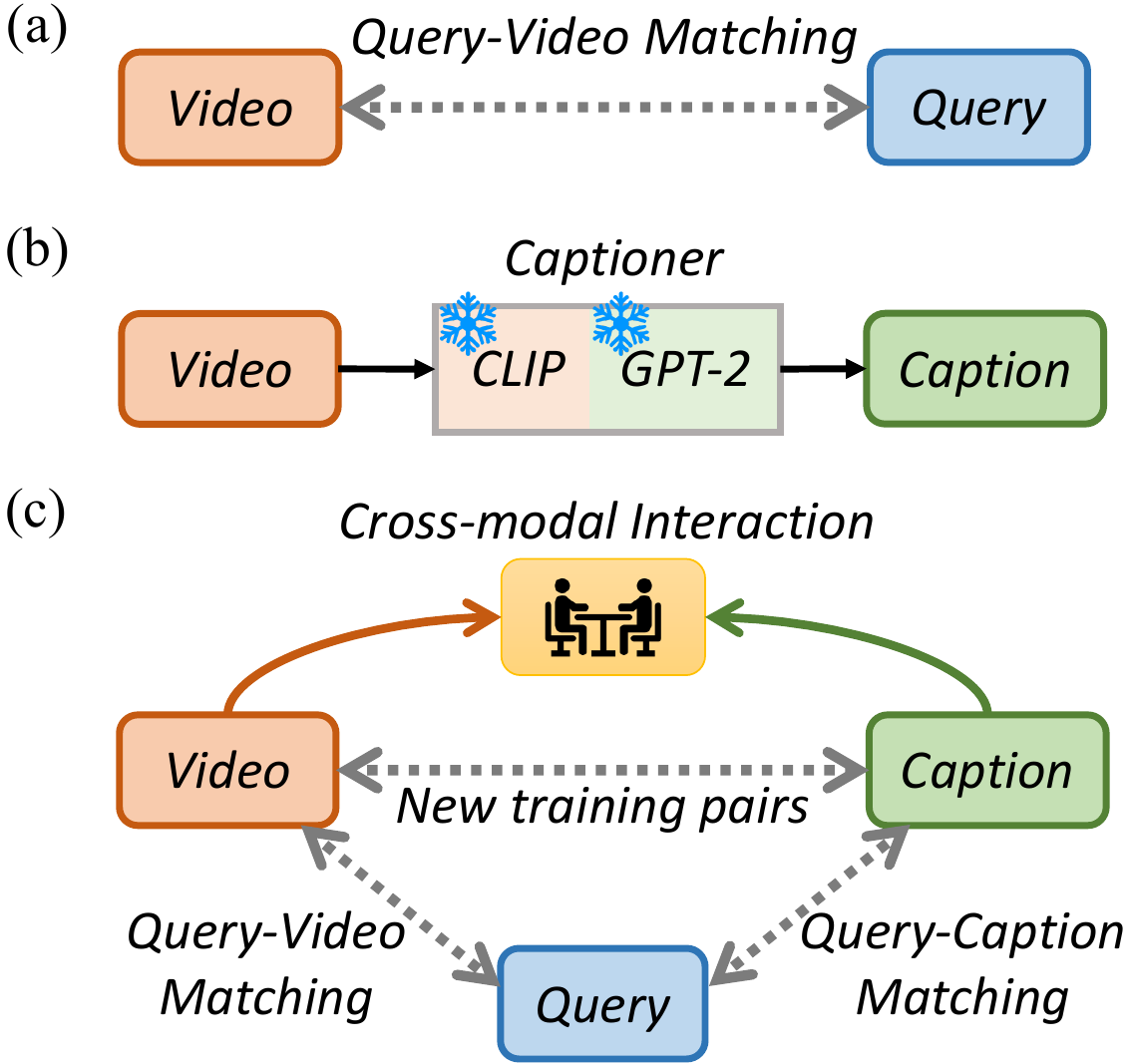}
\end{center}
\caption{(a) An existing end-to-end learning paradigm for text-video retrieval.  (b) Zero-shot video captioning achieved by guiding a large language model (LLM) such as GPT-2~\cite{GPT} with CLIP~\cite{clip}. (c) Our Cap4Video framework leverages the generated captions in three aspects: input data augmentation, intermediate feature interaction, and output score fusion.
}
\label{fig:overview}
\end{figure}


Text-video retrieval is a fundamental task in video-language learning. 
With the rapid advancements in image-language pre-training~\cite{clip,ALIGN,yu2022coca,yuan2021florence}, researchers have focused on expanding pre-trained image-language models, especially CLIP~\cite{clip}, to tackle the text-video retrieval task. The research path has evolved from the most direct global matching (\ie, video-sentence alignment~\cite{luo2022clip4clip,gao2021clip2tv}) to fine-grained matching (\eg, frame-word alignment~\cite{wang2022disentangled}, video-word alignment~\cite{gorti2022xpool}, multi-hierarchical alignment~\cite{fang2021clip2video,min2022hunyuan_tvr}, \etc).
These studies have demonstrated remarkable performance and significantly outperformed previous models.
Two key factors contribute to this improvement. Firstly, CLIP offers powerful visual and textual representations that are pre-aligned in the semantic embedding space, thereby reducing the challenge of cross-modal learning in video-text matching. Secondly, these methods can fine-tune the pre-trained vision and text encoders using sparsely sampled frames in an end-to-end manner.
All of these methods aim to learn cross-modal alignment between the visual representation of videos and the textual representation of the corresponding query, as depicted in Figure~\ref{fig:overview}(a).

However, in real-life scenarios, online videos usually come with related content such as the video's title or tag on the video website. In addition to the visual signal in the video, the associated textual information can also be used to some extent to describe the video content and match the query (\ie, the common text-to-text retrieval). This raises a pertinent question: \emph{How can we generate associated text descriptions for videos?}
One possible solution is to crawl the video title from the video website. However, this method relies on annotations, and there is a risk that the video URL may have become invalid. Another automated solution is to generate captions using zero-shot video caption models. Therefore, we turn our attention to knowledge-rich pre-trained models to handle such challenging open-set scenarios.
We find that the recent study ZeroCap~\cite{tewel2022zerocap} provides a good practice to use frozen CLIP~\cite{clip} and GPT-2~\cite{GPT} for zero-shot image captioning. Thus, we leverage a video extension~\cite{videocap} of ZeroCap for generating captions in the video domain without any further training.

When provided with auxiliary captions, a natural question naturally arises: \emph{
How can we leverage these captions to enhance the text-video retrieval task?} In this paper, we propose the \textbf{Cap4Video} learning framework, as illustrated in Figure~\ref{fig:overview}(c), which utilizes captions in three key ways:
(i) \emph{Input Data}: One simple approach is to augment the training data with the generated captions. Specifically, the given video and its generated caption can be treated as a matched pair, which serves as an additional positive sample pair for training beyond the query-video pairs.
(ii) \emph{Intermediate Feature Interaction}: Cross-modal interaction between the video and captions can be leveraged to improve the video representation. Specifically, we can exploit the complementary information between videos and captions to reduce redundant features from videos and learn more discriminative video representations.
(iii) \emph{Output score}: The generated caption can also represent the video's content, allowing us to employ query-caption matching to complement standard query-video matching for the text-video retrieval task. Moreover, a two-stream architecture can be utilized to reduce model bias and produce more robust results.

We hope that our novel paradigm will encourage further investigation into the video-language learning.
In summary, our contributions are as follows:
\begin{itemize}
    \item We explore a novel problem: leveraging auxiliary captions to further enhance existing text-video retrieval. Besides labor-intensive manual crawling of video website titles, we investigate the potential of rich captions automatically generated by large language models (LLMs) to benefit text-video retrieval.
    \item We propose the \textbf{Cap4Video} learning framework, which maximizes the utility of the auxiliary captions through three aspects: input data, feature interaction, and output score. Our framework improves the performance of existing query-video matching mechanisms, including global matching and fine-grained matching.
    \item Extensive experiments conducted on four video benchmarks demonstrate the effectiveness of our method. Our Cap4Video achieves state-of-the-art performance on MSR-VTT~\cite{xu2016msrvtt} (51.4\%), VATEX~\cite{wang2019vatex} (66.6\%), MSVD~\cite{wu2017msvd} (51.8\%), and DiDeMo~\cite{anne2017didemo} (52.0\%).
\end{itemize}

\section{Methodology}

\begin{figure*}[t]
\begin{center}
\includegraphics[width=1\linewidth]{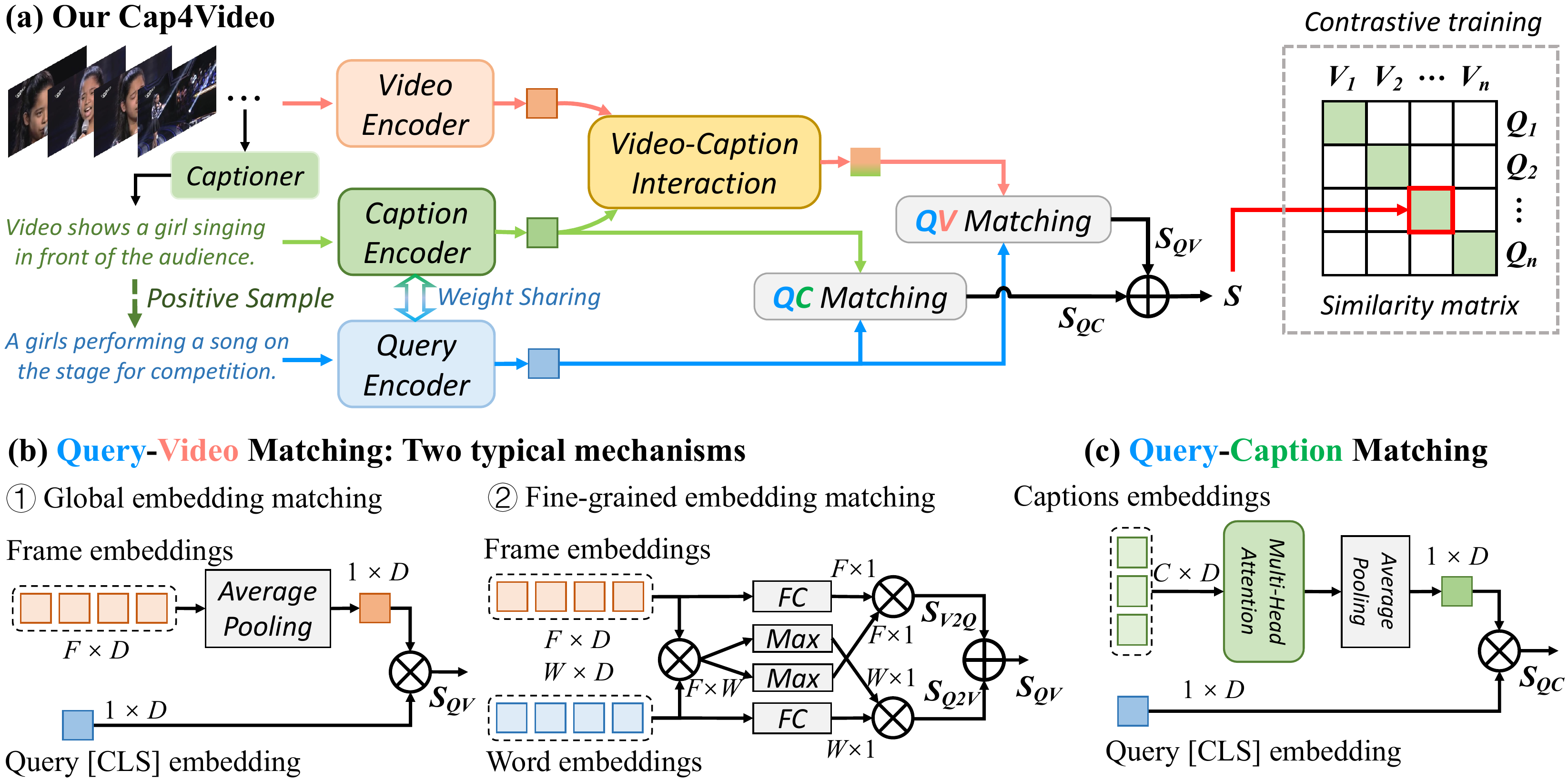}
\end{center}
\vspace{-0.5em}
\caption{An overview of our \textbf{Cap4Video} for text-video retrieval. We first generate captions using a zero-shot video captioner that combines CLIP~\cite{clip} with GPT-2~\cite{GPT}, leveraging knowledge from both frozen web-scale models. We then utilize the pre-extracted caption information from three different perspectives:
i) \emph{Input data:} We use the video and captions to create new positive pairs for data augmentation during training.
ii) \emph{Feature interaction:} We perform feature interaction between video and caption to capture intra- and inter-modality context, yielding enhanced video representations.
iii) \emph{Output score:} The Query-Caption matching branch can complement the original Query-Video matching branch for text-video retrieval.}
\label{fig:approach}
\end{figure*}

\subsection{Background: Text-Video Matching}
Text-video matching aims to evaluate the similarity between a given sentence $Q_i$ and a given video $V_j$, typically using a similarity function $s(Q_i, V_j)$. In text-to-video retrieval, the goal is to rank all videos based on their similarity scores to a given query sentence.
To improve text-video retrieval, recent works~\cite{luo2022clip4clip,gao2021clip2tv,fang2021clip2video} have applied CLIP~\cite{clip} for initialization, leveraging pre-trained knowledge from image-text learning.
Our baselines for text-video matching include two typical mechanisms: global matching and fine-grained matching, as illustrated in Figure~\ref{fig:approach}(b).

\textbf{Global Matching} is a commonly used technique in cross-modal contrastive learning~\cite{clip, jia2021ALIGN, luo2022clip4clip}. In global matching, each modality is encoded independently to obtain global features, which are then used to calculate similarity. We train the visual encoder to output $F$ frame embeddings for a given video that samples $F$ frames. Similarly, the query encoder returns $W$ word embeddings and the [CLS] embedding as the global representation for a given query sentence that contains $W$ words. The frame embeddings are integrated using average pooling to obtain the global video embedding, which is then compared with the global query embedding to calculate similarity.

\textbf{Fine-grained Matching} focuses on modeling the token-level alignment between two modalities, such as frame-word alignment. In order to achieve token-level patch-word alignment for image-text learning, FILIP~\cite{yao2021filip} and ColBERT~\cite{khattab2020colbert} employ a \emph{Max-Mean} pipeline. This pipeline finds the token-wise maximum similarity between patch and word tokens, and then averages the maximum similarity of tokens in the image or text to obtain the similarity between an image and a text or vice versa. Moreover, DRL~\cite{wang2022disentangled} extends the token-wise alignment to text-video retrieval and introduces an attention mechanism to learn weighted pooling instead of mean pooling. We have adopted this mechanism as our enhanced baseline.

\subsection{Preprocessing: Caption Generation}
To obtain auxiliary captions for a given video, we consider the following two approaches.

\noindent\textbf{Manual Crawling of Video Titles.} We extract the video website title by crawling the original links (such as YouTube ID) of each video and utilize it as the caption. However, we skip this step for videos with expired links.

\noindent\textbf{Automatic Video Captioning.} In contrast to the manual approach that relies on annotations, we leverage knowledge from the LLM to generate rich and diverse captions. Given the scalability of our framework, we aim to generate captions directly from downstream videos without any additional training, a process known as zero-shot video captioning. 
To achieve this, we follow \cite{tewel2022zerocap,videocap} and use GPT-2~\cite{GPT} to predict the next word from an initial prompt, \eg, ``Video shows''. A calibrated CLIP~\cite{clip} loss is then used to drive the model to generate sentences that describe the video, incorporating video-related knowledge into the auto-regressive process.
\emph{See Supplementary for more details.}

\subsection{Data Augmentation with Auxiliary Captions}
\label{sec:aug}
Auxiliary captions can be used to augment training data. For example, for a dataset consisting of $N$ videos and their corresponding query sentences, each video and its generated caption can be considered as a positive sample pair for training, in addition to the original query-video pairs. By selecting one caption per video, we can add at least $N$ pairs as additional data augmentation during training.

The automatic video captioner can generate multiple captions (\eg, 20) for each video. However, some of these captions may contain noise and may not be entirely relevant to the video content. To avoid negative effects on training, we use a filtering mechanism that evaluates the semantic similarity between each caption and the ground-truth query of the video using a pre-trained text encoder. The caption with the highest similarity is then chosen for data augmentation.
Note that we only use the ground-truth query for caption filtering during the training phase.

\subsection{Video-Caption Cross-Modal Interaction}\label{sec:interaction}
We further consider taking advantage of the complementarity between videos and captions to reduce redundant features and learn more discriminative video representations.
To preserve the pre-trained CLIP encoder architecture for efficient transfer learning, we limit the interaction to the final caption and frame embeddings. 
Specifically, we pass the frame embeddings $\mathbf{e_v} = \{\mathbf{v}_1,\mathbf{v}_2,\cdots,\mathbf{v}_F\}$ and caption embeddings $\mathbf{e_c} = \{\mathbf{c}_1,\mathbf{c}_2,\cdots,\mathbf{c}_C\}$ to the interaction module, where $F$ and $C$ represent the number of frames and captions, respectively. Figure~\ref{fig:approach}(b) depicts several ways of interaction between the two modalities.

\textbf{Sum.} To obtain an enhanced frame embedding, an intuitive approach is to compute the sum of the global caption embedding $\mathbf{c_g}$ and each frame embedding:
\begin{equation}
\mathrm{Sum}(\mathbf{v}_i, \mathbf{c_g}) = \mathbf{v}_i + \mathbf{c_g}, \qquad  i = 1, \cdots, F,
\end{equation}
where $\mathbf{v}_i \in \mathbb{R}^{D}$ is the $i$-th frame embedding, and $\mathbf{c_g} \in \mathbb{R}^{D}$ is computed by averaging the [CLS] embeddings of $C$ generated captions: 
$\mathbf{c_g} = \frac{1}{C} \sum_{i=1}^C \mathbf{c}_i$.

\textbf{MLP.} To model weighted combinations of each frame embedding and the global caption embedding $\mathbf{c_g}$, we concatenate them together and pass the result through a learnable Multi-layer Perceptron (MLP):
\begin{equation}
    \mathrm{MLP}(\mathbf{v}_i, \mathbf{c_g}) = f_\theta([\mathbf{v}_i, \mathbf{c_g}]), \quad \text{for} \,\, i = 1, \cdots, F,
\end{equation}
where $[\cdot,\cdot]$ denotes the concatenation operation, $f_\theta$ is the MLP with parameter $\theta$.

\textbf{Cross Transformer.} We also investigate the use of self-attention~\cite{vaswani2017attention} for interactions. The Cross Transformer operates on a sequence $\{\mathbf{e_v}, \mathbf{e_c}\} = \{\mathbf{v}_1,\cdots,\mathbf{v}_F,\mathbf{c_1},\cdots,\mathbf{c}_C\}$ and processes them through $L$ encoder-style transformer blocks to generate final representations:
\begin{equation}
    \mathrm{Cross}(\mathbf{e_v}, \mathbf{e_c}) = f_{\psi}(\{\mathbf{e_v}, \mathbf{e_c}\}),
\end{equation}
where $\{\}$ denotes that $\mathbf{e_v}$ and $\mathbf{e_c}$ form a sequence, and $f_{\psi}$ represents the transformer encoders with parameter $\psi$.

\textbf{Co-attention Transformer.} Co-attention~\cite{lu2019vilbert} is another common method for exchanging information between modalities, allowing for mutual attention between video and caption. After this co-attentional transformer layer, we include $L$ transformer layers to model temporal information:


\begin{equation}
    \mathrm{CoAttn}(\mathbf{e_v}, \mathbf{e_c}) = f_{\phi_2}(f_{\phi_1}(\{\mathbf{e_v}, \mathbf{e_c}\})),
\end{equation}
where $f_{\phi_1}$ is the co-attentional transformer with parameter $\phi_1$ and $f_{\phi_2}$ is the transformer encoders with parameter $\phi_2$.

The video-caption interaction module generates frame embeddings that can be further processed based on the type of matching needed. For global matching, the frame embeddings can be averaged to obtain a single video representation. Alternatively, for fine-grained matching, the individual frame embeddings can be retained.

\begin{figure}[t]
\begin{center}
\includegraphics[width=1\linewidth]{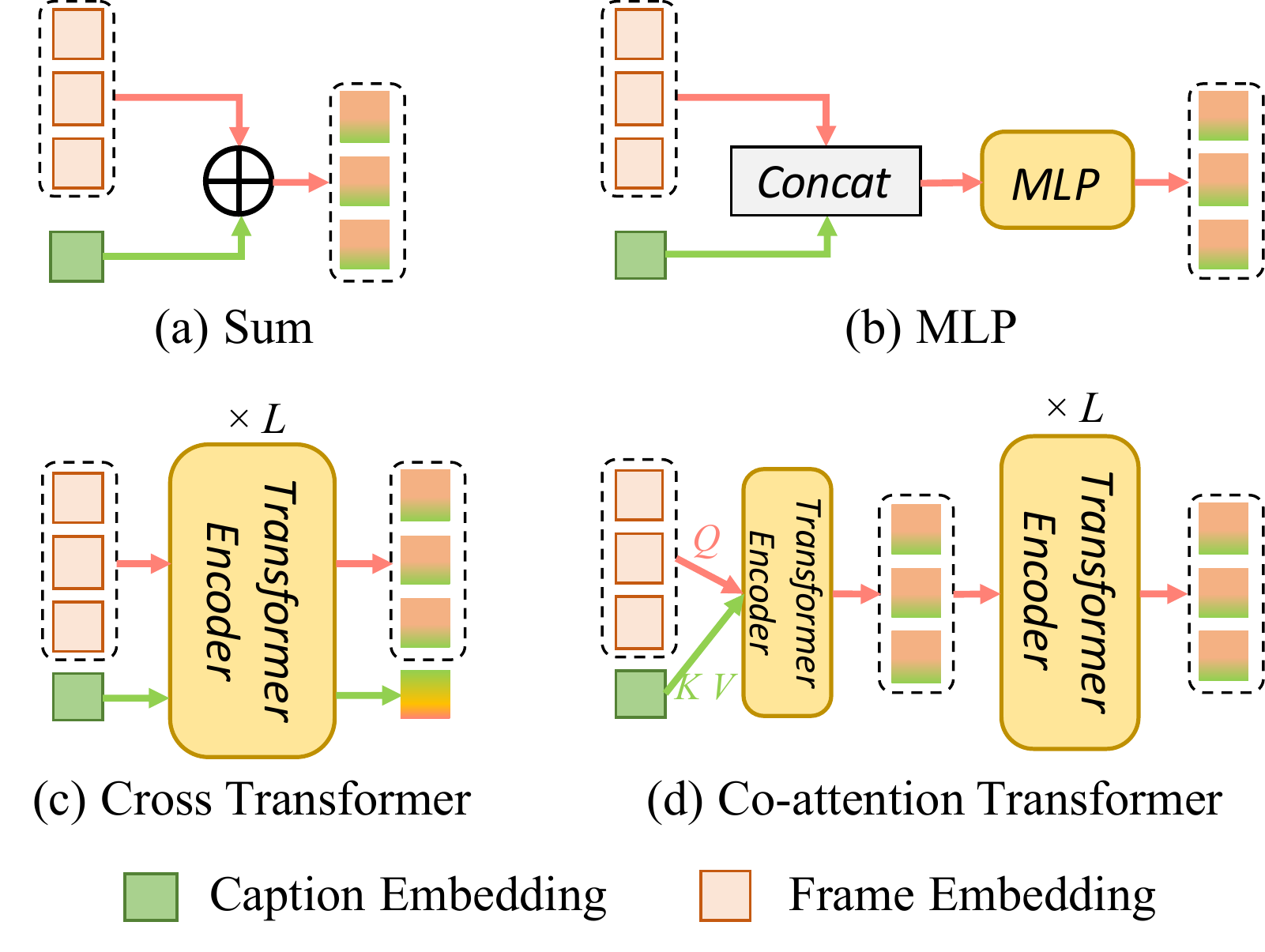}
\end{center}
\vspace{-1em}
\caption{Illustration of four Video-Caption interaction strategies. The enhanced frame embeddings will be followed by a mean pooling for global matching or will remain for fine-grained matching.}
\label{fig:interaction}
\end{figure}

\subsection{Complementary Query-Caption Matching}
Besides using the caption for data augmentation and video feature enhancement, it can also directly represent the video content, allowing for text-text retrieval. 
Specifically, each of the $C$ captions generated by the video is then passed through the caption encoder to obtain its [CLS] text embedding. These caption embeddings are then aggregated to form a global representation, as illustrated in Figure~\ref{fig:approach}(c). The cosine similarity between this global caption embedding and the global query embedding is then calculated to complement the query-video matching.

\textbf{Notation.} Let $\{\mathbf{e_v}_i, \mathbf{e_t}_i, \mathbf{e_c}_i \}_{i=1}^{B}$ be a batch of $B$ triples, where $\mathbf{e_v}_i$, $\mathbf{e_t}_i$, and $\mathbf{e_c}_i$ denote the $i$-th video, query, and caption embedding, respectively. Note that the term ``embedding'' used here is more general for convenience and can vary in meaning depending on the situation. For instance, in query-video global matching, $\mathbf{e_v}_i$ and $\mathbf{e_t}_i$ represent the averaged video feature and global [CLS] text feature, respectively. In query-video fine-grained matching, $\mathbf{e_v}_i$ and $\mathbf{e_t}_i$ represent a sequence of frame embeddings and a sequence of word embeddings, respectively. In query-caption matching, $\mathbf{e_c}_i$ represents a sequence of caption embeddings, and $\mathbf{e_t}_i$ represents a global [CLS] text feature.

\textbf{Learning Objectives.} For the \emph{Query-Caption} branch, we want the caption embedding $\mathbf{e_c}$ and the query embedding $\mathbf{e_t}$ to be close while they are related and far apart when they are not during training phase. 
We follow the common practice \cite{luo2022clip4clip,wang2022disentangled} to consider the bidrectional learning objective. We employ symmetric cross-entropy loss to maximize the similarity between matched \emph{Query-Caption} pairs and minimize the similarity for other pairs: 
\begin{equation}
\begin{gathered}
\mathcal{L}_{Q2C} =  - \frac{1}{B} \sum_{i}^{B}
\log \frac{ \exp(s_{qc}(\mathbf{e_t}_i,\mathbf{e_c}_i)/\tau)  }{\sum_{j}^{B}  \exp(s_{qc}(\mathbf{e_t}_i,\mathbf{e_c}_j)/\tau) }, \\
\mathcal{L}_{C2Q}	= - \frac{1}{B} \sum_{i}^{B} 
\log \frac{ \exp(s_{qc}(\mathbf{e_t}_i,\mathbf{e_c}_i)/\tau)  }{\sum_{j}^{B}  \exp(s_{qc}(\mathbf{e_t}_j,\mathbf{e_c}_i)/\tau) }, \\
\mathcal{L}_{QC} = \frac{1}{2} (\mathcal{L}_{Q2C} + \mathcal{L}_{C2Q}),
\end{gathered}    
\end{equation}
where $s_{qc}(\cdot,\cdot)$ represents the query-caption matching similarity function shown in Figure~\ref{fig:approach}(c), and $\tau$ refers to the temperature hyper-parameter for scaling. 
Similarly, the contrastive loss for \emph{Query-Video} branch is formulated as:
\begin{equation}
\begin{gathered}
\mathcal{L}_{Q2V} =  - \frac{1}{B} \sum_{i}^{B}
\log \frac{ \exp(s_{qv}(\mathbf{e_t}_i,\mathbf{e_v}_i)/\tau)  }{\sum_{j}^{B}  \exp(s_{qv}(\mathbf{e_t}_i,\mathbf{e_v}_j)/\tau) }, \\
\mathcal{L}_{V2Q}	= - \frac{1}{B} \sum_{i}^{B} 
\log \frac{ \exp(s_{qv}(\mathbf{e_t}_i,\mathbf{e_v}_i)/\tau)  }{\sum_{j}^{B}  \exp(s_{qv}(\mathbf{e_t}_j,\mathbf{e_v}_i)/\tau) }, \\
\mathcal{L}_{QV} = \frac{1}{2} (\mathcal{L}_{Q2V} + \mathcal{L}_{V2Q}),
\end{gathered}    
\end{equation}
where $s_{qv}(\cdot,\cdot)$ represents the query-video matching (\eg, global matching, fine-grained matching) similarity function shown in Figure~\ref{fig:approach}(b).
The total loss $\mathcal{L}$ is the sum of \emph{Query-Video} loss $\mathcal{L}_{QV}$ and \emph{Query-Caption} loss $\mathcal{L}_{QC}$:
\begin{equation}
    \mathcal{L} = \mathcal{L}_{QV} + \mathcal{L}_{QC}.
\end{equation}


\section{Experiments: Text-Video Retrieval}
\subsection{Setups}
\noindent\textbf{Datasets.} We conduct experiment on four popular benchmarks for video-to-text retrieval and text-to-video retrieval tasks.
MSR-VTT~\cite{xu2016msrvtt} contains a total of 10K video clips, each having 20 captions. Following the data splits from ~\cite{gabeur2020mmt, miech2019howto100m, luo2022clip4clip}, we train models with associated captions on the \texttt{Training-9K} set and report results on the \texttt{test 1K-A} set. 
DiDeMo~\cite{anne2017didemo} has 10K videos paired with 40K descriptions. Following previous works~\cite{luo2022clip4clip, bain2021frozen, lei2021clipbert}, we concatenate all descriptions of one video to a single query, acting as a \textit{video-paragraph} retrieval task.
VATEX~\cite{wang2019vatex} collects $\sim$35K videos, each with multiple annotations. There are $\sim$26K videos for training, 1,500 videos for validation and 1,500 videos for testing.
MSVD~\cite{wu2017msvd} contains 1,970 videos with 80K captions, with $\sim$40 captions on average per video. There are 1,200, 100, and 670 videos in the train, validation, and test sets, respectively.

\noindent\textbf{Evaluation Metrics.} 
For brevity, we abbreviate Recall at $K$ to R@$K$ ($K=1,5,10$) upon all datasets, which computes the percentage of correct videos among the top $K$ retrieved videos given textual queries (Text$\rightarrow$Video, and vice versa). 
MdR, Median Rank, computes the median of the ground-truth in the retrieval ranking list. 
MnR, Mean Rank, computes the mean rank of the correct results in the retrieval ranking list. 
Note that for MdR and MnR, the lower score means the better (indicated as $\textcolor{black}{\downarrow}$).

\noindent\textbf{Implementation Details.} 
All experiments use the visual encoder in CLIP~\cite{clip} as the video encoder, and the textual encoder in CLIP as both the caption encoder and query encoder. The caption encoder and query encoder share parameters.
To reduce conflict between the two branches, the query-video branch is trained first, followed by the query-caption branch. The text length is fixed to 32, and the video length is fixed to 12 for all datasets except DiDeMo (64 max words and 64 frames).
The initial learning rate is set to 1e-7 for the clip parameters and 1e-4 for the non-clip parameters. The model is trained with a batch size of 128 for 5 epochs, except for DiDeMo (15 epochs), using the Adam~\cite{kingma2015adam} optimizer. All learning rates follow the cosine schedule with a linear warmup~\cite{goyal2017warmup} strategy. 
For the number of generated captions per video, we set $C$ to 30. The interaction module employs $L$ transformer layers, where $L$ is set to 4 for VATEX and MSR-VTT, and 1 for Didemo and MSVD. In the caption branch, the number of transformer layers is set to 2.
For caption generation, we directly use the original pre-trained CLIP and GPT-2, without any additional tuning.

\subsection{Comparison with State-of-the-Arts}
In this section, we compare our Cap4Video with recent state-of-the-art methods on four benchmarks: MSR-VTT~\cite{xu2016msrvtt}, MSVD~\cite{wu2017msvd}, VATEX~\cite{wang2019vatex}, and DiDeMo~\cite{anne2017didemo}.

Table~\ref{tab:sota-didemo} shows the comparisons on DiDeMo, where our Cap4Video outperforms CLIP4Clip~\cite{luo2022clip4clip} by a significant margin of \textbf{9.2\%} in R@1 and exceeds DRL~\cite{wang2022disentangled} by \textbf{3.0\%}, demonstrating the effectiveness of our method.

Table~\ref{tab:sota-MSRVTT} provides a comparison of our approach with recent state-of-the-art models on MSR-VTT. Our method achieves new state-of-the-art performance on text-to-video retrieval for both ViT-B/32 and ViT-B/16 backbones, significantly surpassing previous works. For instance, we achieve a \textbf{+4.8\%} higher R@1 than CLIP4Clip with the same ViT-B/32 on text-to-video retrieval. Additionally, our Cap4Video outperforms the recent TS2-Net~\cite{liu2022ts2net} by \textbf{2.3\%} and \textbf{2.0\%} with ViT-B/32 and ViT-B/16, respectively.

Table~\ref{tab:sota-msvd} and Table~\ref{tab:sota-vatex} show the results for the MSVD and VATEX datasets, respectively, where we use ViT-B/16 as our backbone. For MSVD, our Cap4Video achieves a remarkable performance of 51.8\% R@1 and outperforms CLIP-based models CLIP4Clip~\cite{luo2022clip4clip} and X-Pool~\cite{ma2022xclip} by \textbf{6.6\%} and \textbf{4.6\%} on text-to-video retrieval, respectively. For VATEX, our approach also outperforms the recent state-of-the-art methods and achieves a \textbf{+7.5\%} R@1 improvement over TS2-Net~\cite{liu2022ts2net} for text-to-video retrieval.

\begin{table}[b]
  \centering
    \setlength\tabcolsep{4pt}
    \scalebox{0.95}{
    \begin{tabular}{lccccc}
    \toprule
    \textbf{Method} & \textbf{R@1} & \textbf{R@5} & \textbf{R@10} & \textbf{MdR} & \textbf{MnR} \\
    \midrule
    CE~\cite{liu2019CE} & 15.6 & 40.9 & - & 8.2 & - \\
    ClipBERT~\cite{lei2021clipbert} & 21.1 & 47.3 & 61.1 & 6.3 & - \\
    Frozen~\cite{bain2021frozen} & 31.0 & 59.8 & 72.4 & 3.0 & - \\
    TMVM~\cite{lin2022textadaptive} & 36.5 & 64.9 & 75.4 & 3.0 & - \\
    CLIP4Clip~\cite{luo2022clip4clip} & 42.8 & 68.5 & 79.2 & 2.0 & 18.9 \\
    TS2-Net~\cite{liu2022ts2net} & 41.8 & 71.6 & 82.0 & 2.0 & 14.8 \\
    HunYuan~\cite{min2022hunyuan_tvr} & 45.0 & 75.6 & 83.4 & 2.0 & 12.0 \\
    DRL~\cite{wang2022disentangled} & 49.0 & 76.5 & 84.5 & 2.0 & - \\
    \midrule
    \rowcolor{gray!20} \textbf{Cap4Video}  & \textbf{52.0} & 79.4 & 87.5 & 1 & 10.5 \\
    \bottomrule
  \end{tabular}}
  \vspace{-1mm}
  \caption{Results of text-to-video retrieval on the DiDeMo~\cite{anne2017didemo}.}
  \label{tab:sota-didemo}
\end{table}

\begin{table*}
  \centering
   \scalebox{0.95}{
    \begin{tabular}{lc|ccccc|ccccc}
    \toprule
    \multirow{2}{*}{\textbf{Method}} & \multirow{2}{*}{\textbf{Venue}} & \multicolumn{5}{c}{\textbf{Text $\rightarrow$ Video}} & \multicolumn{5}{c}{\textbf{Video $\rightarrow$ Text}} \\
       & & \textbf{R@1} & \textbf{R@5} & \textbf{R@10} & \textbf{MdR$\textcolor{black}{\downarrow}$} & \textbf{MnR$\textcolor{black}{\downarrow}$} & \textbf{R@1} & \textbf{R@5} & \textbf{R@10} & \textbf{MdR$\textcolor{black}{\downarrow}$} & \textbf{MnR$\textcolor{black}{\downarrow}$} \\
    \midrule
    ClipBERT~\cite{lei2021clipbert} & CVPR'20 & 22.0 & 46.8 & 59.9 & 6.0 & - & - & - & - & - & \\
    MMT~\cite{gabeur2020mmt} & ECCV'20 & 26.6 & 57.1 & 69.6 & 4.0 & - & 27.0 & 57.5 & 69.7 & 3.7 & 21.3 \\
    T2VLAD~\cite{wang2021t2vlad} & CVPR'21 & 29.5 & 59.0 & 70.1 & 4.0 & - & 31.8 & 60.0 & 71.1 & 3.0 \\
    SupportSet~\cite{patrick2020supportset} & ICLR'21 & 30.1 & 58.5 & 69.3 & 3.0 & - & 28.5 & 58.6 & 71.6 & 3.0 & -\\
    Frozen~\cite{bain2021frozen} &  ICCV'21 & 32.5 & 61.5 & 71.2 & 3.0 & - & - & - & - & - & - \\
    BridgeFormer~\cite{ge2022bridgeformer} & CVPR'22 & 37.6 & 64.8 & 75.1 & - & - & - & - & - & - & - \\
    TMVM~\cite{lin2022textadaptive} & NeurIPS'22 & 36.2 & 64.2 & 75.7 & 3.0 & - & 34.8 & 63.8 & 73.7 & 3.0 & - \\
    \midrule
    \rowcolor{mygray}
    \multicolumn{12}{l}{\emph{CLIP-ViT-B/32}} \\
    CLIP4Clip~\cite{luo2022clip4clip} & arXiv'21 & 44.5 & 71.4 & 81.6 & 2.0 & 15.3 & 42.7 & 70.9 & 80.6 & 2.0 & 11.6 \\
    CenterCLIP~\cite{zhao2022centerclip} & SIGIR'22 & 44.2 & 71.6 & 82.1 & 2.0 & 15.1 & 42.8 & 71.7 & 82.2 & 2.0 & 10.9 \\
    CAMoE~\cite{cheng202camoe} & arXiv'21 & 44.6 & 72.6 & 81.8 & 2.0 & 13.3 & 45.1 & 72.4 & 83.1 & 2.0 & 10.0 \\
    CLIP2Video~\cite{fang2021clip2video} & arXiv'21 & 45.6 & 72.6 & 81.7 & 2.0 & 14.6 & 43.5 & 72.3 & 82.1 & 2.0 & 10.2 \\
    X-Pool~\cite{gorti2022xpool} & CVPR'22 & 46.9 & 72.8 & 82.2 & 2.0 & 14.3 & - & - & - & - & - \\
    QB-Norm~\cite{bogolin2022qbnorm} & CVPR'22 & 47.2 & 73.0 & 83.0 & 2.0 & - & - & - & - & - & - \\
    TS2-Net~\cite{liu2022ts2net} & ECCV'22 & 47.0 & 74.5 & 83.8 & 2.0 & 13.0 & 45.3 & 74.1 & 83.7 & 2.0 & 9.2 \\
    DRL~\cite{wang2022disentangled} & arXiv'22 & 47.4 & 74.6 & 83.8 & 2.0 & - & 45.3 & 73.9 & 83.3 & 2.0 & - \\
    \rowcolor{gray!20} \textbf{Cap4Video}  & & \textbf{49.3} & 74.3 & 83.8 & 2.0 & 12.0 & \textbf{47.1} & 73.7 & 84.3 & 2.0 & 8.7 \\
    \midrule
    \rowcolor{mygray}
    \multicolumn{12}{l}{\emph{CLIP-ViT-B/16}} \\
    CLIP2TV~\cite{gao2021clip2tv} & arXiv'21 & 48.3 & 74.6 & 82.8 & 2.0 & 14.9 & 46.5 & 75.4 & 84.9 & 2.0 & 10.2 \\
    CenterCLIP~\cite{zhao2022centerclip} & SIGIR'22 & 48.4 & 73.8 & 82.0 & 2.0 & 13.8 & 47.7 & 75.0 & 83.3 & 2.0 & 10.2 \\
    TS2-Net~\cite{liu2022ts2net} & ECCV'22 & 49.4 & 75.6 & 85.3 & 2.0 & 13.5 & 46.6 & 75.9 & 84.9 & 2.0 & 8.9 \\
    DRL~\cite{wang2022disentangled} & arXiv'22 & 50.2 & 76.5 & 84.7 & 1.0 & - & 48.9 & 76.3 & 85.4 & 2.0 & - \\
    \rowcolor{gray!20} \textbf{Cap4Video}  & & \textbf{51.4} & 75.7 &  83.9 & 1.0 & 12.4 & \textbf{49.0} & 75.2 & 85.0 & 2.0 & 8 \\
    \bottomrule
  \end{tabular}}
  \vspace{-1mm}
  \caption{Retrieval results on the validation set of MSR-VTT 1K~\cite{xu2016msrvtt}. Here we report results \textbf{without} any post-processing operations (\eg, DSL~\cite{cheng202camoe} or QB-Norm~\cite{bogolin2022qbnorm}) during inference.
  }
  \label{tab:sota-MSRVTT}
\end{table*}

\begin{table}[t]
  \centering
    \setlength\tabcolsep{4pt}
    \scalebox{0.95}{
    \begin{tabular}{lccccc}
        \toprule
        \textbf{Method} & \textbf{R@1} & \textbf{R@5} & \textbf{R@10} & \textbf{MdR} & \textbf{MnR} \\
        \midrule
        CE~\cite{liu2019CE} & 19.8 & 49.0 & 63.8 & 6.0 & - \\
        SUPPORT~\cite{patrick2020supportset} & 28.4 & 60.0 & 72.9 & 4.0 & - \\
        CLIP~\cite{clip} & 37.0 & 64.1 & 73.8 & 3.0 & - \\
        Frozen~\cite{bain2021frozen} & 33.7 & 64.7 & 76.3 & 3.0 & - \\
        TMVM~\cite{lin2022textadaptive} & 36.7 & 67.4 & 81.3 & 2.5 & - \\
        CLIP4Clip~\cite{luo2022clip4clip} & 45.2 & 75.5 & 84.3 & 2.0 & 10.3 \\
        X-Pool~\cite{gorti2022xpool} & 47.2 & 77.4 & 86.0 & 2.0 & 9.3 \\
        \midrule
    \rowcolor{gray!20} \textbf{Cap4Video}  & \textbf{51.8} & 80.8 & 88.3 & 1 & 8.3 \\ 
        \bottomrule
      \end{tabular}}
      \vspace{-1mm}
  \caption{Results of text-to-video retrieval on the MSVD~\cite{wu2017msvd}.}
  \label{tab:sota-msvd}
\end{table}

\begin{table}
  \centering
  \setlength\tabcolsep{4pt}
  \scalebox{0.95}{
    \begin{tabular}{lccccc}
        \toprule
        \textbf{Method} & \textbf{R@1} & \textbf{R@5} & \textbf{R@10} & \textbf{MdR} & \textbf{MnR} \\
        \midrule
        HGR~\cite{chen2020HGR} & 35.1 & 73.5 & 83.5 & 2.0 & - \\
        CLIP~\cite{clip} & 39.7 & 72.3 & 82.2 & 2.0 & 12.8 \\
        SUPPORT~\cite{patrick2020supportset} & 44.9 & 82.1 & 89.7 & 1.0 & - \\
        CLIP4Clip~\cite{luo2022clip4clip} & 55.9 & 89.2 & 95.0 & 1.0 & 3.9 \\
        Clip2Video~\cite{fang2021clip2video} & 57.3 & 90.0 & 95.5 & 1.0 & 3.6 \\
        QB-Norm~\cite{bogolin2022qbnorm} & 58.8 & 88.3 & 93.8 & 1.0 & - \\
        TS2-Net~\cite{liu2022ts2net} & 59.1 & 90.0 & 95.2 & 1.0 & 3.5 \\
        \midrule
    \rowcolor{gray!20} \textbf{Cap4Video}  & \textbf{66.6} & 93.1 & 97.0 & 1 & 2.7 \\ 
        \bottomrule
  \end{tabular}}
  \vspace{-1mm}
  \caption{Results of text-to-video retrieval on the VATEX~\cite{wang2019vatex}.}
  \label{tab:sota-vatex}
\end{table}

\begin{table*}[t]
  \centering
  \scalebox{0.95}{
    \begin{tabular*}{\linewidth}{@{}@{\extracolsep{\fill}}l|ccccc|ccccc@{}}
    \toprule
    \multirow{2}{*}{\textbf{Method}} & \multicolumn{5}{c}{\textbf{Global Matching}} & \multicolumn{5}{c}{\textbf{Fine-grained Matching}} \\
      & \textbf{R@1} & \textbf{R@5} & \textbf{R@10} & \textbf{MdR$\textcolor{black}{\downarrow}$} & \textbf{MnR$\textcolor{black}{\downarrow}$} & \textbf{R@1} & \textbf{R@5} & \textbf{R@10} & \textbf{MdR$\textcolor{black}{\downarrow}$} & \textbf{MnR$\textcolor{black}{\downarrow}$} \\
    \midrule
Baseline & 42.8 & 70.4 & 79.0 & 2 & 16.6 & 45.7 & 73.7 & 82.6 & 2 & 13.1 \\
    \hline
     \rowcolor{gray!20}
    \multicolumn{11}{l}{\emph{+Different Sources of Caption as Data Augmentation}} \\
     Video Title from Source URL  & 43.8 & 71.1 & 80.9 & 2 & 15.1 & 44.3 & 72.7 & 83.5 & 2 & 13.1 \\
     Zero-shot Video Captioning  & \textbf{44.2} & 70.7 & 81.5 & 2 & 16.2 & \textbf{46.3} & 72.5 & 81.7 & 2 & 12.9 \\ 
     \hline
     \rowcolor{gray!20}
    \multicolumn{11}{l}{\emph{+Different Number of Captions for Data Augmentation}} \\
     Top-1 & \textbf{44.2} & 70.7 & 81.5 & 2 & 16.2 & \textbf{46.3} & 72.5 & 81.7 & 2 & 12.9 \\
     Top-3  & 43.3 & 71.7 & 81.6 & 2 & 15.0 & 45.5 & 73.8 & 82.4 & 2 & 12.7 \\ 
     Top-5  & 43.4 & 70.6 & 80.4 & 2 & 16.2 & 45.6 & 72.7 & 82.7 & 2 & 12.9 \\ 
     \hline     
     \rowcolor{gray!20}
    \multicolumn{11}{l}{\emph{+Video-Caption Feature Interaction}} \\ 
    Video Only & 44.2 & 70.7 & 81.5 & 2 & 16.2 & 46.3 & 72.5 & 81.7 & 2 & 12.9 \\
    Sum & 43.8 & 71.5 & 80.3 & 2 & 16.1 & 47.2 & 73.3 & 82.8 & 2 & 13.1 \\
    Concat-MLP & 37.5 & 66.1 & 78.4 & 3 & 15.7 & 40.0 & 68.7 & 79.9 & 2 & 12.7 \\
    Cross Transformer & 44.6 & 71.6 & 80.3 & 2 & 14.6 & 47.9 & 75.4 & 83.0 & 2 & 11.5 \\
    Co-attention Transformer & \textbf{45.3} & 71.2 & 80.9 & 2 & 15.0 & \textbf{48.5} & 74.0 & 82.5 & 2 & 12.7 \\
    \hline
     \rowcolor{gray!20}
    \multicolumn{11}{l}{\emph{+Query-Caption Matching Score}} \\     
    Query-Video Only & 45.3 & 71.2 & 80.9 & 2 & 15.0 & 48.5 & 74.0 & 82.5 & 2 & 12.7 \\
    Query-Caption Only & 30.3 & 55.2 & 67.5 & 4 & 26.4 & 30.3 & 55.2 & 67.5 & 4 & 26.4  \\
    Query-Video + Query-Caption & \textbf{45.6} & 71.7 & 81.2 & 2 & 14.8 & \textbf{49.3} & 74.2 & 83.4 & 2 & 12.1 \\
    \bottomrule
  \end{tabular*}}
  \vspace{-1mm}
  \caption{Component-wise evaluation of our framework on the MSR-VTT 1K validation set. With the ViT-B/32 backbone, we report the text-to-video retrieval results for two representative Query-Video matching mechanisms: global matching and fine-grained matching. The consistent improvement on two typical matching mechanisms demonstrates the generalization ability and effectiveness of our method. }
  \label{tab:main-ablation}
\end{table*}

In recent studies, several methods have been proposed to improve text-video retrieval performance by adjusting similarity during inference using other query information. Notably, our results adheres to the standard retrieval logic, where the most relevant video is retrieved for each query from a set of videos, without any knowledge of the relationship between other queries and videos. Therefore, all results reported in our tables do not involve any post-processing procedures such as DSL~\cite{cheng202camoe} and QB-Norm~\cite{bogolin2022qbnorm}.
Overall, the consistent state-of-the-art performance across four benchmarks demonstrates the effectiveness of our Cap4Video.

\subsection{Ablation Study}
In this section, we provide detailed ablation studies to clarify the effects of each part of our design. 

\noindent\textbf{Auxiliary Caption as Data Augmentation.}
We begin by investigating the impact of captions on data augmentation for training.
In a real-world scenario, the original video title would naturally serve as an additional auxiliary caption. Therefore, we manually extracted the title from the video's original webpage and compared it to the caption generated by the GPT-2 model.
Table~\ref{tab:main-ablation} presents the results of using different sources of the caption, from which we observe that using captions generated by the web-scale model as data augmentation for training can lead to direct improvements in R@1 (+1.4\%, +0.6\%) under both matching mechanisms. Using video titles can also bring a 1\% improvement for global matching.

We also explore the impact of the number of generated captions used for augmentation. We used the caption filtering mechanism mentioned in \cref{sec:aug} to rank the relevance of captions and the ground-truth query, and selected different numbers of captions for training. The results demonstrated that using only one caption is sufficient.

\noindent\textbf{Benefit on Both Online and Offline Videos Scenarios.}
Our method is applicable for both online and offline videos. Offline videos are local videos with no title, while online videos have a title on the video website. Therefore, for online videos, the step of generating a caption with CLIP+GPT-2 is skipped, and the website title is used directly as a caption. In Table~\ref{table:cap}, the results show that our method has significantly improved over the global matching baseline for \textbf{both} offline and online videos.

\noindent\textbf{Video-Caption Feature Interaction.}
As mentioned in \cref{sec:interaction}, we have designed four approaches for Video-Caption feature interaction. Based on the results presented in Table~\ref{tab:main-ablation}, we can conclude the following:
1) The basic approach of \emph{Sum} has been shown to enhance fine-grained matching by 0.9\% R@1, but there is no noticeable improvement in global matching.
2) The \emph{MLP} approach is difficult to optimize and performs poorly in both matching scenarios. We speculate that the \emph{MLP}'s operation in a black-box environment, despite creating a nonlinear metric space, may lead to degradation.
3) The \emph{Cross Transformer} approach has demonstrated improvements of +0.4\% and +1.6\% in two matching settings, respectively. These enhancements may be attributed to the self-attention mechanism's ability to capture the inter-modal relationship between the video and caption.
4) Moreover, the \emph{Co-attention Transformer} approach has significantly boosted performance, with gains of +1.1\% and +2.2\% for these two matching mechanisms.
In summary, the results indicate that proper interaction between the video and generated caption can lead to better video representation and improved Query-Video matching.

\begin{table}[t]
\centering
\scalebox{0.95}{
\begin{tabular}{ccl}
\toprule
  & \textbf{Caption Source} & \textbf{R@1} \\
\midrule
Baseline & N/A & 42.8 \\ \midrule
\multirow{2}{*}{Cap4Video} & Original Website Title & 45.8 \\
& Captioner (CLIP+GPT-2) & 45.6 \\
\bottomrule
\end{tabular}}
\vspace{-1mm}
\caption{Exploring the effectiveness of captions from different sources on MSR-VTT 1k-A. Setting: ViT-B/32, global matching.}
\label{table:cap}
\end{table}

\noindent\textbf{Query-Caption matching.} 
We also investigate the Query-Caption matching branch for text-video retrieval. We aggregate caption embeddings using mean pooling to yield a global embedding. As shown in Table~\ref{tab:main-ablation}, the single Query-Caption matching branch achieves 30.3\% R@1 on text-to-video retrieval, outperforming previous query-video matching methods such as ClipBERT~\cite{lei2021clipbert} (22.0\%) and MMT~\cite{gabeur2020mmt} (26.6\%). This suggests that the Query-Caption matching branch can complement the regular Query-Video matching branch for improved text-video retrieval. Combining the score of Query-Caption matching branch with Query-Video matching branch further improves performance (+0.8\%).

Overall, Cap4Video utilizes generated captions in three ways: input data augmentation, intermediate feature interaction, and output score fusion, leading to consistent improvements (+2.8\% / +3.6\%) in both matching mechanisms.

\begin{figure}[t]
\begin{center}
\includegraphics[width=1\linewidth]{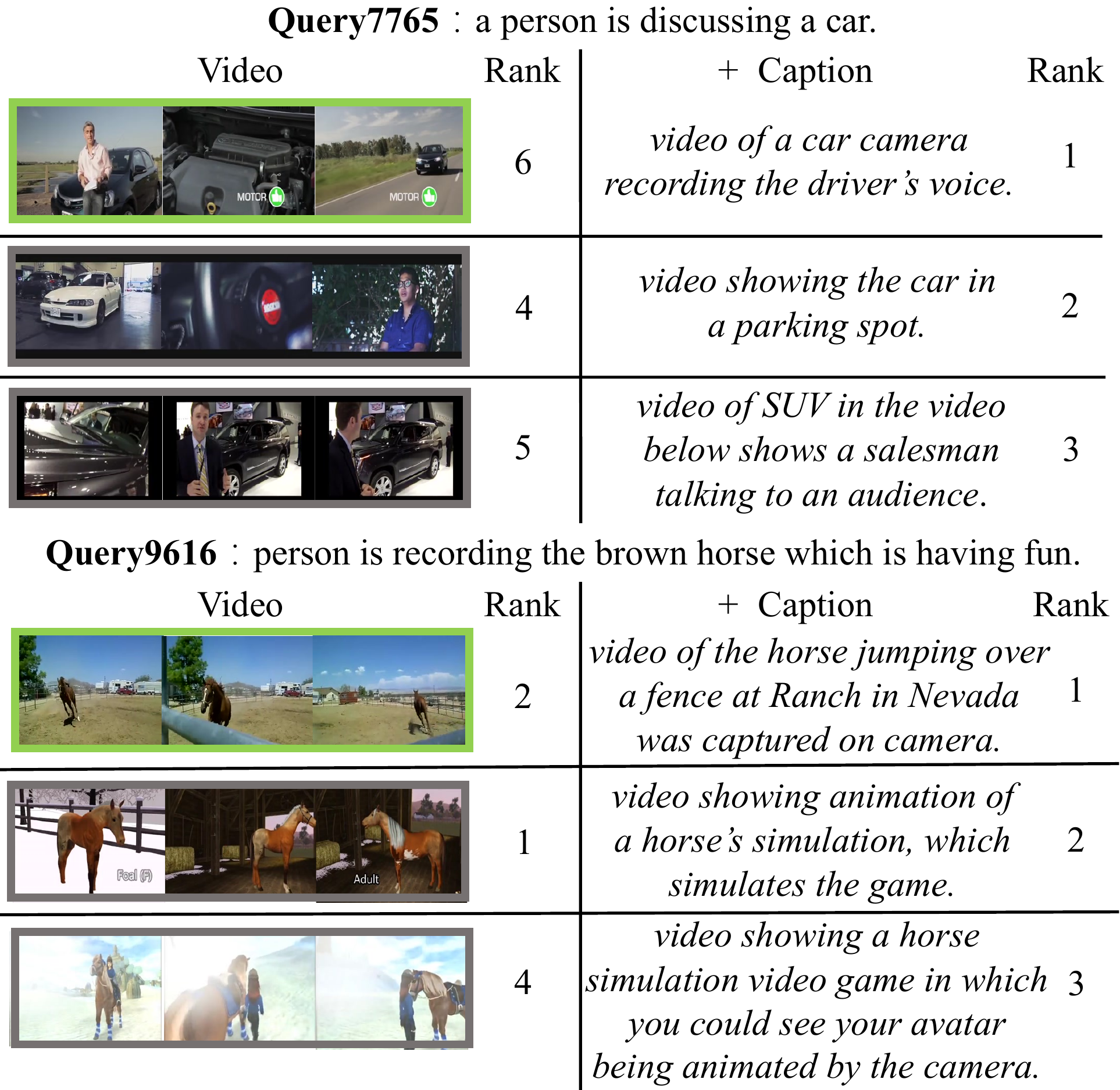}
\end{center}
\vspace{-1em}
\caption{The text-video results on the MSR-VTT 1K-A test set. \textbf{Left:} The ranking results of the query-video matching model. \textbf{Right:} The ranking results of Cap4Video, which incorporates generated captions to enhance retrieval. Please zoom in for best view.
}
\label{fig:vis}
\end{figure}

\subsection{Visualization}
We provide two examples of videos retrieved by our Cap4Video method and a model without auxiliary captions. As illustrated in Figure~\ref{fig:vis}, our approach successfully retrieves the ground-truth video with the assistance of the caption, while the video-only model returns multiple videos that are somewhat relevant to the query but not precise. \emph{See more qualitative results in Supplementary}.

\section{Related Works}
\noindent\textbf{Zero-shot Image Captioning.}
In the field of natural language processing, transformer-based GPT models~\cite{GPT,GPT3} have been successful in generating text from prompts by training on large-scale text corpora. Similarly, CLIP~\cite{clip}, a vision-language alignment model trained on 400 million image-text pairs, has demonstrated impressive zero-shot performance on vision tasks. However, research on transferring web-scale models to zero-shot video captioning remains limited.
Recently, ZeroCap~\cite{tewel2022zerocap} proposed a method of using CLIP and the GPT-2 language model to generate textual descriptions of input images, leveraging knowledge from both models in a truly zero-shot manner without re-training or fine-tuning model parameters. MAGIC~\cite{su2022magic} has also used CLIP scores to align GPT-2 logits with corresponding images but requires fine-tuning on the MS-COCO caption text corpus.
More recently, a study~\cite{videocap} extended the zero-shot capability of ZeroCap to the video domain. In this paper, we employ this video extension to generate auxiliary captions without any additional training.

\noindent\textbf{Text-Video Retrieval}
 aims to retrieve relevant video content based on natural language descriptions. 
Early studies~\cite{gabeur2020mmt, liu2019CE, wang2021DMM, chen2020HGR, wang2021t2vlad} focused on knowledge transfer from ``expert" models and captured intra-modal and cross-modal interactions based on pre-extracted features. However, the performance of these methods is limited since they cannot perform end-to-end optimization.
Recently, more methods have involved end-to-end training for text-video retrieval. One typical approach~\cite{miech2019howto100m, bain2021frozen, miech2020MIL-NCE} is to first perform large-scale text-video pre-training, then transfer the model to downstream text-video retrieval tasks. 
With the emergence of pre-trained Vision-Language Models (VLMs), there have been increased efforts to leverage them for improving video understanding~\cite{bike, text4vis}.
Thus, another training-efficient line is to directly expand the pre-trained VLM to the text-video retrieval task. CLIPBERT~\cite{lei2021clipbert} enables affordable pioneering end-to-end training with a sparse sampling strategy. After that, recent works~\cite{luo2022clip4clip, bogolin2022qbnorm, zhao2022centerclip, gorti2022xpool, liu2022ts2net, gao2021clip2tv, fang2021clip2video, wang2022align, bain2022clip, fang2023uatvr} focus on transferring knowledge from CLIP models that have been pre-trained on 400M image-text pairs. The research path has evolved from the most direct global matching (\ie, video-sentence alignment~\cite{luo2022clip4clip, gao2021clip2tv}) to fine-grained matching (\eg, frame-word alignment~\cite{wang2022disentangled}, video-word alignment~\cite{gorti2022xpool}, multi-hierarchical alignment~\cite{fang2021clip2video, min2022hunyuan_tvr}).
Unlike above CLIP-Based efforts on query-video matching, we propose to generate auxiliary captions from videos to improve text-video retrieval. Thus our method is compatible with both global and fine-grained matching.

\section{Conclusion}

We introduce Cap4Video, a novel framework that leverages captions generated by web-scale language models to enhance text-video matching in three key ways: 1) Input data augmentation for training, 2) Intermediate video-caption feature interaction for compact video representations, and 3) Output score fusion for improved text-video retrieval.
Our approach demonstrates consistent performance gains on four standard text-video retrieval benchmarks, outperforming state-of-the-art methods by a clear margin.

{\small
\bibliographystyle{ieee_fullname}
\bibliography{egbib}
}

\clearpage
\twocolumn[\begin{center}\textbf{\Large Cap4Video: What Can Auxiliary Captions Do for Text-Video Retrieval? \vspace{2mm}
\\ \textnormal{\emph{Supplementary Material}}}\end{center}
\vspace{1em}
\begin{center}
\large Wenhao Wu$^{1,2}$\qquad
Haipeng Luo$^{3}$\qquad
Bo Fang$^{3}$\qquad
Jingdong Wang$^{2}$\qquad
Wanli Ouyang$^{4,1}$\\
$^1$The University of Sydney \qquad $^2$Baidu Inc. \\ 
$^3$University of Chinese Academy of Sciences  \qquad $^4$Shanghai AI Laboratory\\
{\tt\small whwu.ucas@gmail.com}
\vspace{2.5em}
\end{center}]

\appendix
\setcounter{table}{0}
\setcounter{figure}{0}
\renewcommand{\thetable}{A.\arabic{table}}
\renewcommand{\thefigure}{A.\arabic{figure}}

In this appendix, 
\S\ref{supp:details} contains \textit{details} of zero-shot video captioner.
\S\ref{supp:video} contains further \textit{results}: computation efficiency (\S\ref{sec:com_eff}), more baselines (\S\ref{sec:cap_agg}), and more visualizations (\S\ref{sec:vis}).

\section{Caption Generation}\label{supp:details}
To obtain auxiliary captions for a given video, we consider the following two approaches.

\textbf{Crawling Titles.} We extract the video website title by crawling the original links (such as YouTube ID) of each video and use it as the caption. For instance, for the MSR-VTT dataset, we crawl the title of the video website as the caption based on the original link provided by the dataset annotation. However, we found that 2555 out of the 10,000 videos in the dataset have invalid links, so we do not use the title as extra auxiliary information in these videos, and only perform video-query matching.

\textbf{Video Captioning.} We utilize the video extension~\cite{videocap} of ZeroCap~\cite{tewel2022zerocap} for zero-shot video captioning. 
In Cap4Video, the captioner can be replaced with other methods if desired.

\begin{figure}[t]
\begin{center}
\includegraphics[width=0.98\columnwidth]{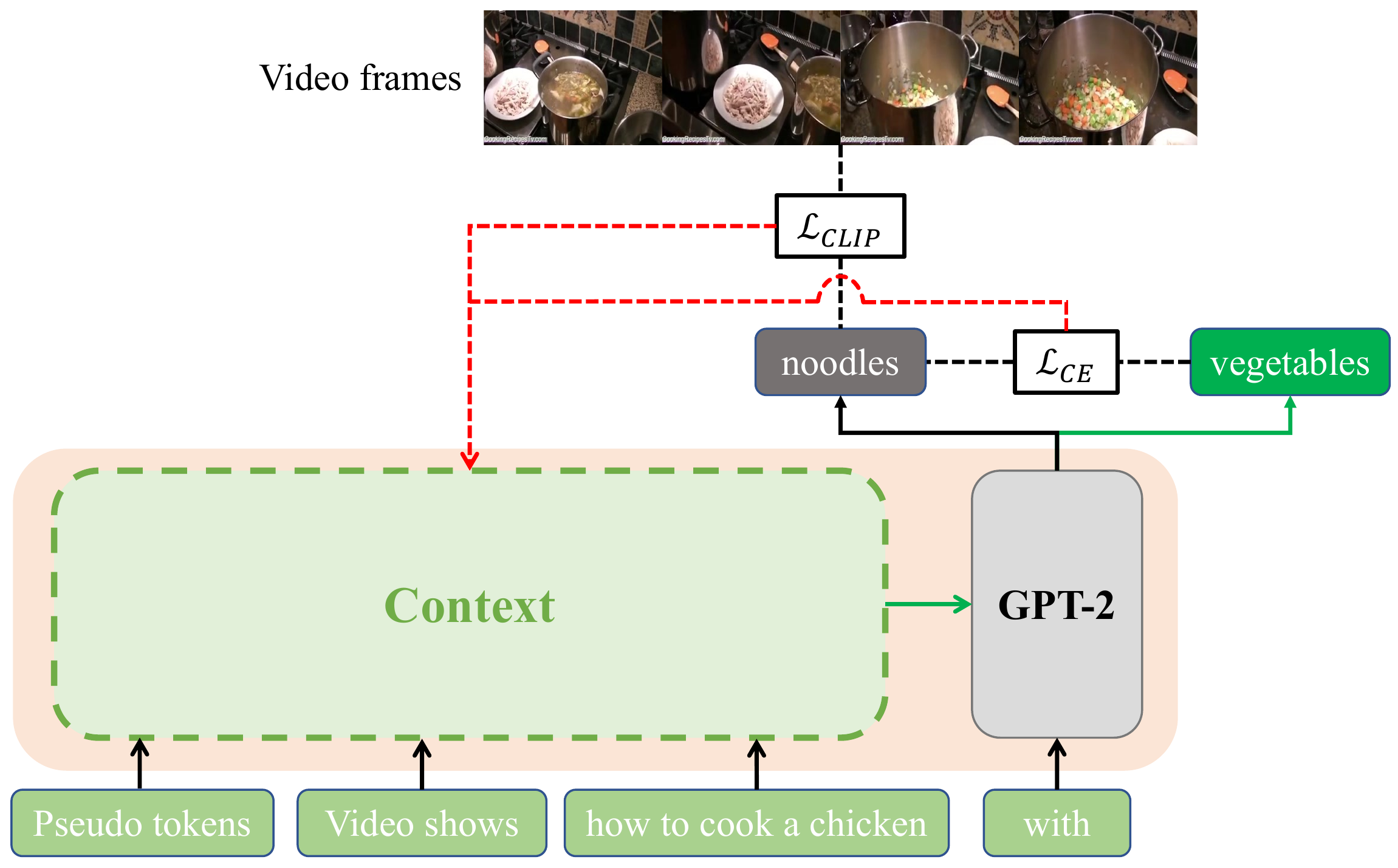}
\end{center}
\caption{Zero-shot video captioning~\cite{videocap} using CLIP and GPT.
}
\label{fig:zerocap}
\end{figure}

ZeroCap employs GPT-2~\cite{GPT}, a transformer-based pre-trained language model, to predict the next word from an initial prompt, such as ``Video shows". To integrate vision-related knowledge into the auto-regression process, the model is motivated to generate sentences that describe a given video using a calibrated CLIP loss $\mathcal{L}_{CLIP}$. An additional loss term, $\mathcal{L}_{CE}$, is employed to keep the next token distribution consistent with the original language model. Optimization occurs during auto-regression, and the process is repeated for each token. Simple arithmetic of visual cues in CLIP's embedding space can capture semantic relations.
Although ZeroCap is effective in describing individual visual cues, it faces a challenge in generating coherent descriptions of multiple images. In contrast to the original ZeroCap approach, the video extension \cite{videocap} optimizes pseudo-tokens through iterative sentence generation, with the goal of steering the overall sentence generation process towards a coherent description of the video, as depicted in Figure~\ref{fig:zerocap}.

In each generation step, the primary aim is to guide GPT-2 towards a desired visual direction. This guidance has two objectives: (i) aligning with the provided video, and (ii) maintaining language attributes. To achieve the first objective, CLIP~\cite{clip} is utilized to assess the similarity of a token to a video and adjust the model's cache accordingly. For the second objective, the objective is regularized to resemble the original target output before modification. The solved optimization problem updates the context cache at each time point. 

As a result, the captioner can generate captions for videos directly without any additional training.
To prevent the generation of long and repetitive sentences, we set a token limit of 20 for each sentence. We also utilized frame sampling with a frame rate of 3 FPS and performed 30 iterations of generation to obtain 30 captions for each video.

For further information on the implementation, please see the paper~\footnote{https://arxiv.org/pdf/2207.11100.pdf} and official code~\footnote{https://github.com/YoadTew/zero-shot-video-to-text}.

\begin{figure*}
\centering
  \includegraphics[width=0.8\textwidth]{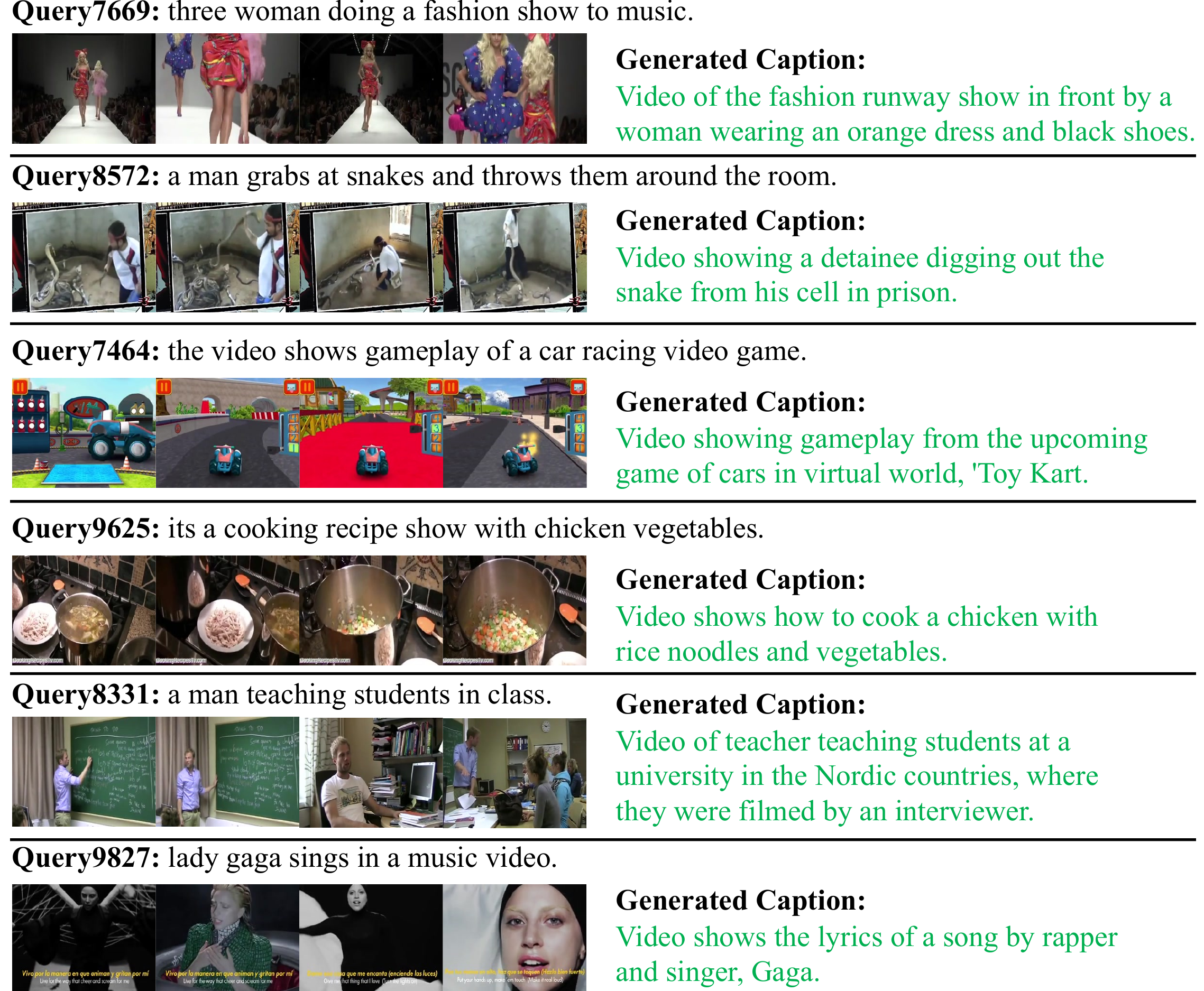} 
  \caption{Examples of auxiliary captions generated by the zero-shot video captioner on the MSRVTT 1K-A test set. These captions help to change the original wrong Top-1 prediction to the correct one.}
  \label{fig:cap}
\end{figure*}

\section{More Results}\label{supp:video}

\subsection{Computation Efficiency}\label{sec:com_eff}
In Table~\ref{table:effi}, we show the computational cost and efficiency. We use single NVIDIA 3090 GPU and a batch size of 16 to measure the throughput.

\begin{table}[t]
\centering
\scalebox{0.91}{
\begin{tabular}{cccc}
\toprule
  & FLOPs & \#Params & Throughput \\
\midrule
Cap4Video & 60.5G & 96.8M &  164.2 vid/s \\ \bottomrule
\end{tabular}}
\vspace{-1mm}
\caption{Computation efficiency. ``vid/s" represents the average number of videos per second. Model: ViT-B/32.}
\label{table:effi}
\end{table}

\subsection{Additional Baselines}\label{sec:cap_agg}
To demonstrate the benefits of GPT-2 and language augmentation, we present the following baselines on the MSR-VTT 1k-A dataset:
1) Ensemble Baseline: we ensemble the zero-shot CLIP score and the finetuned video branch score. The results are shown in Table~\ref{tab:ense}. We can observe that the ensemble score of ``Video+Zero-Shot CLIP" is lower than our Cap4Video (42.9\% vs 43.8\%), demonstrating the advantage of GPT-2.
2) Cap4Video using synthetic captions generated and filtered using the finetuned CLIP model. In our paper, we use the original CLIP and GPT-2 without any fine-tuning to perform zero-shot video captioning on any video. Here we study the effect of finetuned CLIP on the captioner in Table~\ref{tab:caper}. Although the captions generated by finetuned CLIP can bring further improvement, it reduces the method's flexibility.

\begin{table}[t]
\centering
\scalebox{0.91}{
    \begin{tabular}{ccc}
    \toprule
    Video  & +Auxiliary & Ensemble \\ \midrule
    \multirow{2}{*}{42.8} & +Caption & 43.8 \\ 
      & +Zero-Shot CLIP & 42.9 \\
    \bottomrule
    \end{tabular}}
    \vspace{-1mm}
    \caption{Ensemble baselines (ViT-B/32 w/ global matching).
    }
    \label{tab:ense}
\end{table}
\begin{table}[t]
\centering
\scalebox{0.91}{
    \begin{tabular}{cc}
    \toprule
   Captioner use &  Caption \\ \midrule
  Original CLIP & 30.7 \\
   Fine-tuned CLIP &  35.1 \\
    \bottomrule
    \end{tabular}}
    \caption{Caption baseline with different Captioner setting.}
    \label{tab:caper}
\end{table}

\begin{figure*}
\centering
  \includegraphics[width=0.81\textwidth]{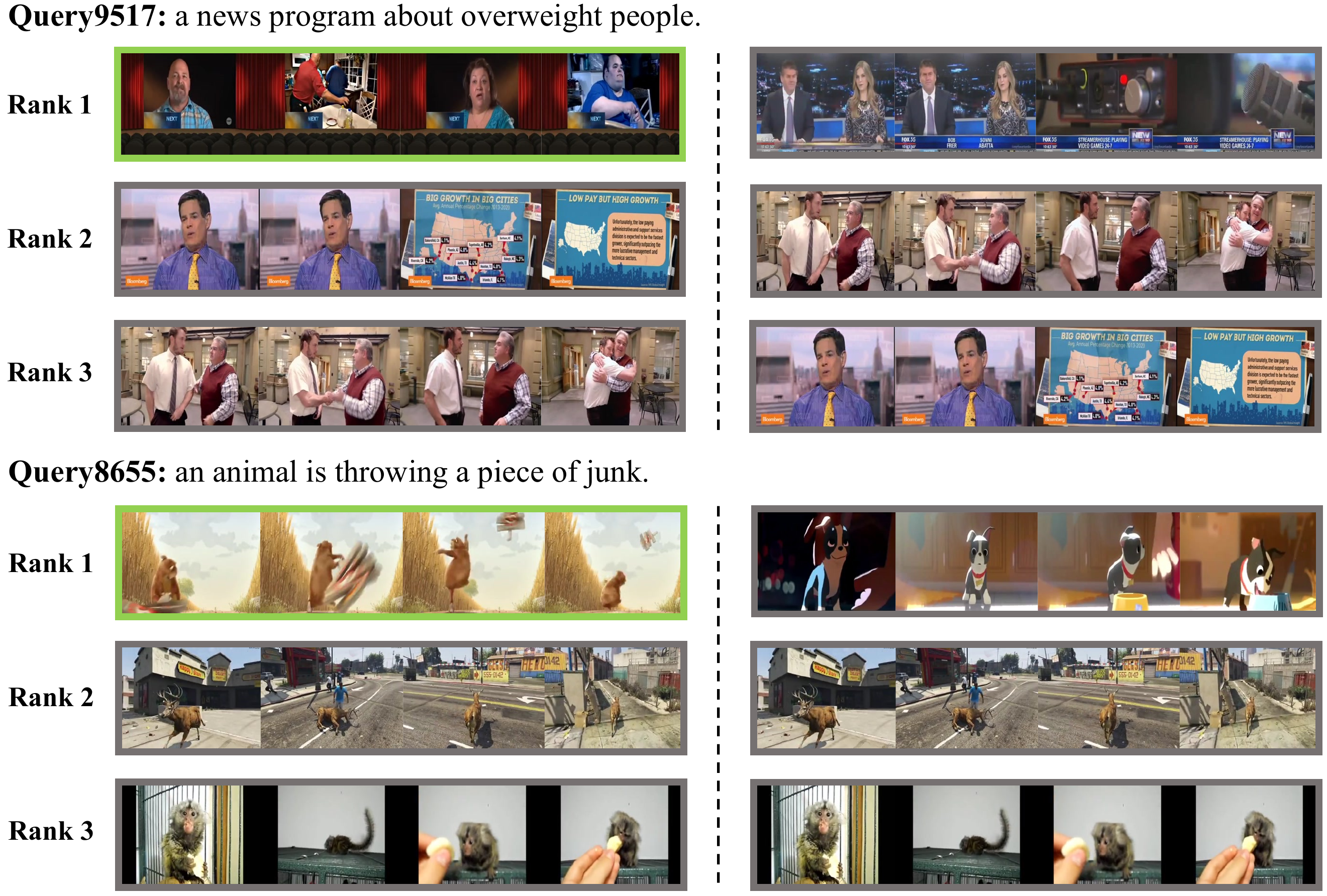} \\[7pt]
  \includegraphics[width=0.81\textwidth]{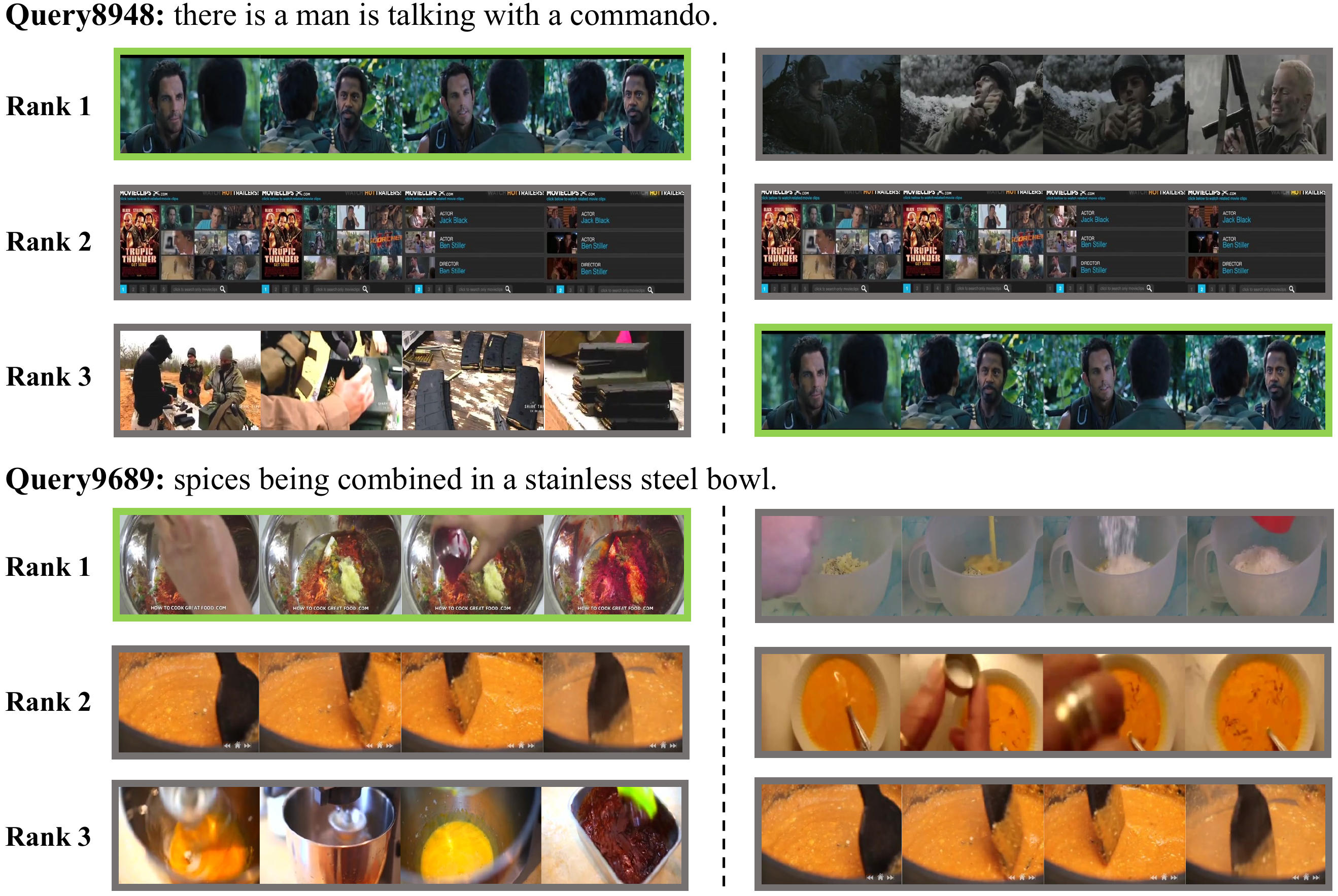}
  \caption{Examples of text-video retrieval results on the MSRVTT 1K-A test set. The left are the videos ranked by our Cap4Video, and the right are the results from the model without involving caption.}
  \label{fig:results}
\end{figure*}

\subsection{Qualitative Results}\label{sec:vis}
To further validate our motivation for using auxiliary caption for text-video retrieval, we present more visualizations of auxiliary captions in Figure~\ref{fig:cap} and retrieval results in Figure~\ref{fig:results}.

\end{document}